\def\year{2021}\relax
\documentclass[letterpaper]{article} % DO NOT CHANGE THIS
\usepackage{aaai21}  % DO NOT CHANGE THIS
\usepackage{times}  % DO NOT CHANGE THIS
\usepackage{helvet} % DO NOT CHANGE THIS
\usepackage{courier}  % DO NOT CHANGE THIS
\usepackage[hyphens]{url}  % DO NOT CHANGE THIS
\usepackage{graphicx} % DO NOT CHANGE THIS
\usepackage{natbib}  % DO NOT CHANGE THIS AND DO NOT ADD ANY OPTIONS TO IT
\usepackage{caption} % DO NOT CHANGE THIS AND DO NOT ADD ANY OPTIONS TO IT
\usepackage{amssymb}% http://ctan.org/pkg/amssymb
\usepackage{pifont}% http://ctan.org/pkg/pifont
\usepackage{enumitem}
\newcommand{\cmark}{\ding{51}}
\newcommand{\xmark}{\ding{55}}

\usepackage{multirow}
\usepackage{makecell}

\title{Toward a Human-Level Video Understanding Intelligence}

\author {
    Yu-Jung Heo\textsuperscript{\rm 1\thanks{These authors contributed equally to this study.}},
    Minsu Lee\textsuperscript{\rm 1,2\footnotemark[1]},
    Seongho Choi \textsuperscript{\rm 1},
    Woo Suk Choi \textsuperscript{\rm 1},
    Minjung Shin \textsuperscript{\rm 3},
    Minjoon Jung \textsuperscript{\rm 1},
    Jeh-Kwang Ryu \textsuperscript{\rm 3} and
    Byoung-Tak Zhang \textsuperscript{\rm 1,2}\\
}

\affiliations {
    \textsuperscript{\rm 1} Seoul National University, \textsuperscript{\rm 2} AI Institute (AIIS), Seoul National University, \textsuperscript{\rm 3} Dongguk University \\
    \{yjheo,mslee,shchoi,wschoi,mjjung,btzhang\}@bi.snu.ac.kr, \{shinymj,ryujk\}@dongguk.edu,
}
\begin{document}
\maketitle

\begin{abstract}
We aim to develop an AI agent that can watch video clips and have a conversation with human about the video story. Developing video understanding intelligence is a significantly challenging task, and evaluation methods for adequately measuring and analyzing the progress of AI agent are lacking as well.
In this paper, we propose the Video Turing Test to provide effective and practical assessments of video understanding intelligence as well as human-likeness evaluation of AI agents. We define a general format and procedure of the Video Turing Test and present a case study to confirm the effectiveness and usefulness of the proposed test.
\end{abstract}

\section{Introduction} % written by mslee
% ultimate goal
AI agents that can watch videos with people and share empathy for video content through various conversations are promising AI applications that people expect. To this end, AI agents have to accurately perceive and recognize contents of a video and have a natural muti-turn conversation with people based on understanding of the contents.
%However, this is currently a very challenging goal. 

% researches on video story intelligence
Recently, researches on text-to-video retrieval, video captioning, and video question answering (videoQA) have been actively conducted to improve video understanding intelligence. In addition, large-scale datasets have been built and publicly available to facilitate the researches~\cite{alamri2019audio,lei2018tvqa,lei2020tvr,choi2021dramaqa}. Studies using these datasets usually apply automatic evaluation metrics to measure the performances of AI agents. 
For videoQA task, multiple-choice QA typically uses overall accuracy, and open-ended QA applies evaluation metrics frequently used in natural language generation tasks (e.g., BLEU~\cite{papineni2002bleu}, METEOR~\cite{banerjee2005meteor}, CIDEr~\cite{vedantam2015cider}).
% ROUGE~\cite{lin2004rouge}

% pros and cons for evaluation metrics -> we adopts Turing Test !
These automatic evaluation metrics are convenient to apply, but they also have limitations. 
% For example, the overall accuracy does not take into account the difficulty of questions or required cognitive components even though it is intuitive and easy to calculate. 
For example, the overall accuracy, while being intuitive and easy to calculate, does not take into account the difficulty of questions or required cognitive components.
In addition, the scores of evaluation metrics for language generation models cannot determine whether the content is a correct answer to the question. 
%For example, the scores of evaluation metrics based on language models cannot determine whether the content is a correct answer to the question. In addition, the overall accuracy does not take into account the difficulty of questions or required cognitive components even though it is intuitive and easy to calculate.
Therefore, these metrics are hard to apply to measure the video understanding ability and the human-likeness of an AI agent.
%Therefore, automatic evaluation metrics for video intelligence cannot correctly measure the human-likeness of an AI agent. 
Furthermore, since story understanding task usually leads to multiple interpretations, it is essential to evaluate how persuasive the answers of AI are for people. 

% Although Turing Test cannot directly measure intelligence, it is suitable for assessing video intelligence based on the validity of answers.

\begin{figure}[t]
\centering
\includegraphics[width=0.9\columnwidth]{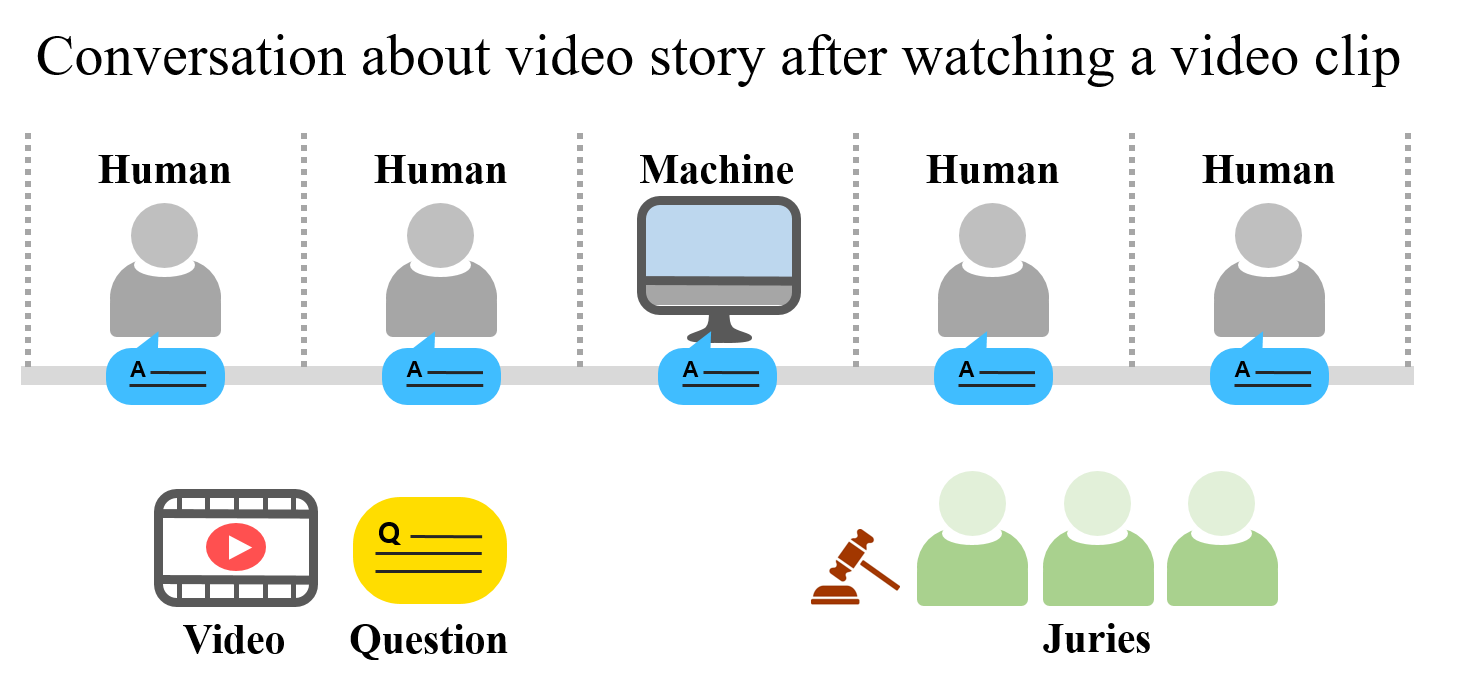}
\caption{An illustration of the Video Turing Test (VTT). Detailed procedure of VTT is described in `Video Turing Test' section.}
\label{fig:vtt}
\end{figure}

% our contributions!
In this paper, we propose a novel Video Turing Test (VTT) as a method for evaluating the video understanding ability and the human-likeness of an AI agent (Figure \ref{fig:vtt}). In VTT, all players see a given video clip consisting of multimodality, then have question and answering about the video story. We also exploit a new evaluation metric, CogME, established on the story elements and thinking strategies. We can analyze intelligence in the aspect of humans' understanding process through the metric.

% \item 객관식: Accuracy % 장: extra annotation 없이 평가 가능, 정답률 기반의 analysis 가능 단: clue가 함께 제공되는 형태, 객관식 형태의 문제를 잘 만들기 어렵다는 점? high cost, well-designed QA 
%\item 주관식: LM metrics % 장: 평가 good, 단: open question이라 평가에 있어서 주관적일 수 있음, extra cost, 분석에 있어서 다소 애매할 수 있음 (여러명의 judge 필요)
% "비디오 이해 인공지능 개발의 목적이 비디오 내용에 대해 대화 나누고 느낌을 공유하는 것이라면, 단순히 accuracy 점수를 높이는 기존의 방식은 단편적인 내용을 (이해 아니고) capture and retrieve 하는 기능에 국한되므로, 기존의 평가 방식은 목적에 부합하지 않다. 게다가 스토리를 이해하는 것은 정답이 없는(? 다양한 해석이 가능한?) 영역이기 때문에, 인공지능의 performance가 얼만큼 다양한 사람들에게 어필(콩글리시인 것 같아요. )하는지를 평가하는 과정이 반드시 필요하다. 튜링 테스트는, 비록 지능을 직접 측정하기 위해 제안된 것이 아닐지라도, 이러한 목적에 매우 부합한다. " 이민수 교수님이 작성하신 intro 내용과 다소 겹치는 경향이 있는데, intro에서 튜링테스트의 유용성을 강하게 어필한 후에 related work으로 넘어가는게 어떨까 합니다.

% Human players and a machine agent answer a given question about a video story. Based on the conversation, juries guess who is the AI agent. 

\section{Related Work}
\subsection{Turing Test}
% Turing test ... 
%Alan Turing, in his seminal paper in 1950, suggested how human-level machine intelligence can be measured~\cite{turing1950computing}. 
%If the machine succeeds more than once in three trials, we can regard the machine passes the Turing test, and the result validates that the machine intelligence achieves human-likeness. 

In 1950, Alan Turing proposed a variation of imitation game, now called Turing Test, where a computational machine and a human player communicate with an interrogator using text while invisible to each other~\cite{turing1950computing}. The objective of the machine is to ensure that the interrogator cannot distinguish the machine from the human player. If the machine achieves the objective, we can regard the machine intelligence can be considered human-like. The Turing Test has been established as a novel measurement, but has been criticized for the following limitations. First, a machine can pass the test based on `trickery or guile'~\cite{weizenbaum1966eliza,shieber1994lessons,boden2006mind}.
Second, the Turing Test focused primarily on language capability without considering additional components of human intelligence (e.g., visual understanding). Third, the passing criteria for the Turing Test, as well as the interrogator's judgement, are quite subjective. 
% for human-likeness in machine intelligence

To complement the limitations, several new suites of tests are suggested by extending the original Turing Test.
% modi qtype, restricted theme of conversation, not only text but multimodal
To provide a more quantitative measure of intelligence preventing passing the test by trickery or guile, \cite{mckinstry1997minimum,geman2015visual} considered only boolean type questions about a series of facts or an image scene, respectively. 
To avoid reliance on conversational ability, \cite{olague2021less} suggested a purely visual processing-based Turing test which aims to imitate the emotional interpretation of humans. On the other hand, \cite{adiwardana2020meena} proposed a novel evaluation metric, Sensibleness and Specificity Average to capture human-likeness of conversational ability of a machine.
%On the other hand, \cite{adiwardana2020meena} proposed a novel evaluation metric, Sensibleness and Specificity Average which captures two aspects (i.e., making sense and being specific) of human-likeness in conversation. Based on the new metric, the human-likeness of machine intelligence can be measured and analyzed.
%\cite{mckinstry1997minimum} introduced Minimum Intelligence Signal Tests consisting only of boolean type questions. The purpose of the test is to provide a more quantitative measure of intelligence preventing passing the test by trickery or guile.
%Specifically, \cite{geman2015visual,olague2021less} introduced the visual Turing test which aims to understand image scenes.
\cite{adams2016athlon} suggested I-athlon as a multidimensional Turing Test. The authors aim to evaluate a wide variety of intelligence behavior such as video understanding, not only the conversational ability of machines. 
Lastly, \cite{zhang2009teaching} proposed a Multimodal Memory Game where human teachers watch a video with an AI agent and teach the agent to generate text from a given image (video frame) or generate an image from a given text. This work shares similar motivations and experimental settings with ours, however, we focus on developing and evaluating the video understanding intelligence via question answering about video story.

\section{Video Turing Test}\label{sec:vtt} % written by mslee
% 목표 / 정의
% VTT 설계 원칙 및 고려사항. 
%VTT를 어떻게 진행하는지 순차적으로 자세하게 서술. video의 content를 대상으로 하기 때문에 AI agent가 단순히 사람을 속이기 위한 속임수를 쓰는 방식으로 해당 test를 통과하긴 어려움
This section describes a novel Turing Test, called Video Turing Test (VTT), for measuring video understanding intelligence. Video understanding ability includes many dimensions of intelligence, such as linguistic ability, visual understanding, story understanding, and reasoning. To circumvent the well-known limitations of the Turing Test described in the previous section, the VTT is designed as a standardized test with an effective and practical assessment of many aspects of video understanding intelligence.
%and a comprehensive measure of AI progress. 

\subsection{Procedure for Video Turing Test}
The general VTT procedure is illustrated in Figure \ref{fig:vtt}. All participants are separated from one another. Especially the jury can not see all players. A set of paired video clips and questions is prepared. Then, the rest of the procedures are as follows: (i) Players and the jury watch a video clip. (ii) A question about the video story is given to both the players and the jury. (iii) Each player submits an answer to the question. (iv) The submitted answers are presented to the jury. (v) After checking all the answers, the jury guesses who the AI agent is and votes for their prediction.
% to the prepared questions

\subsection{Format of Video Turing Test}
Open-ended multi-turn conversations about a given video clip among players and an interrogator may be regarded as the most appropriate approach to confirm the human-likeness of AI agents. However, since the open-ended conversation about the video is analogous to the Turing Test, the resulting conversation scripts naturally inherit the drawbacks of the Turing Test. Therefore, it is difficult to qualitatively and quantitatively measure the degree of video understanding intelligence. 
%Additionally, it is hard to reach a consensus on the appropriate attributes for answers to demonstrate players’ video understanding intelligence. Many people would judge the longer or more detailed answers as to better. However, the answers by human players usually vary significantly. Moreover, it is hard to control the flow of conversation not to go off-topic.
VTT should be easy to evaluate but hard to solve. The test must not be solvable using trickery or guile. To meet these requirements, we argue that a question and answer is more suitable for the VTT. The QA format allows us to thoroughly examine the video understanding intelligence by carefully selecting the set of intended questions and reviewing the players’ answers. Unlike conversation tasks where golden answer does not exist, it is relatively apparent to evaluate answers to questions. 

We can consider three QA types according to the way of expressing the answer to the question. The most definite QA type is to ask multiple-choice questions. Since multiple-choice question requires to pick an answer among several candidate answers, the evaluation process becomes simple. 
%and the players cannot do trickery or cheat. 
%This approach is also suitable to apply to different AI agents to compare their performances.
Another more challenging option is to leave the answer open-ended. For short-answer questions, assessing the correctness of the answers is straightforward. Short-answer questions are frequently used in quiz shows because this type of QA is simple and clear. Full-sentence-answer questions are the most appropriate method for examining human-level video understanding intelligence. The full-sentence answer represents players’ linguistic competence and thinking perspectives as well as correctness of the answer. Juries can get great hints about the human-likeness of players from their fluency. However, in order to evaluate whether each full-sentence answer is correct or not, it requires an additional human scoring work based on the conversational human-likeness (sensibleness) and correctness of the answer (specificity)~\cite{adiwardana2020meena}. 
%Since manual scoring is a very subjective task, inconsistent or contradictory scores among human evaluators frequently occur. Proper guidance or training for evaluators can help to reduce the large variances in scores~\cite{clark2021EvalHumanEval}. 
Organizers of the VTT can select an appropriate QA format according to the circumstances such as the characteristics of the players, the target audience and the test environment.

\begin{table*}[]
\centering
\begin{tabular}{ll}
\Xhline{2\arrayrulewidth}
% \multicolumn{2}{l}{\textbf{Question:} How did Haeyoung feel when Dokyung suddenly shows up?} \\ 
% \multicolumn{2}{l}{\textbf{Story elements:} Character, Event and Emotion in Target, Feature and Sequence in Content, Reasoning in Thinking}
% \\
% \hline
% AI agent & Answer: Haeyoung felt surprised. (\cmark)\\
% Pre-operational stage & Answer: She was going to work and he scared her. (\xmark)\\
% Middle concrete stage & Answer: She was surprised, then she said it was good to see him. (\cmark)\\
% Concrete generalization stage & Answer: She was surprised to see him. (\cmark)\\
% Formal stage & Answer: She is initially startled but afterwards, she is happy to see him. (\cmark)\\
% \hline
% \hline
\multicolumn{2}{l}{\textbf{Question:} What did Dokyung do in this scene?} \\
\multicolumn{2}{l}{\textbf{Story elements:} Character and Event in Target, Identity in Content, Recognition in Thinking} \\
\hline
AI agent & Answer: Dokyung was sitting on the ground. (\xmark)\\
Pre-operational stage & Answer: Dokyung is looking at Haeyoung. (\xmark)\\
Middle concrete stage & Answer: He is sitting down next to Haeyoung. (\cmark)\\
Concrete generalization stage & Answer: He was sat on the hospital bed. (\cmark)\\
Formal stage & Answer: Dokyung was sitting in the bench in the hospital within Haeyoung's sight. (\cmark)\\
\Xhline{2\arrayrulewidth}
\end{tabular}
% \caption{Two question and answer examples sampled in the case study of the Video Turing Test (VTT). The check-mark (\cmark) and x-mark (\xmark) represent correct and wrong answers, respectively.}
\caption{A question and answer example sampled in the case study of the Video Turing Test (VTT). The check-mark (\cmark) and x-mark (\xmark) represent correct and wrong answers, respectively.}
\label{table:qual}
\end{table*}

\subsection{Evaluation metrics for Video Turing Test}
We aim to evaluate whether an AI agent understands video, and to analyze the test results so that future research directions can be suggested based on the current strengths and weaknesses of the AI agent. 
To this end, it is important to prepare carefully selected a set of questions that can evaluate the players’ video understanding ability in a systematic way. 
In addition, by checking the answers to the question, it should be possible to evaluate each player’s video understanding intelligence specifically. 
By using a systematic evaluation metric for composing a set of questions and analyzing the level of intelligence of players, we can improve the reliability and usefulness of the VTT.

For a systematic evaluation of video understanding intelligence, we develop Cognitive Modules for Evaluation (CogME) which is a top-down evaluation system for videoQA based on the cognitive process of human and story elements~\cite{shin2021cogme}.
CogME is composed of three cognitive modules: \texttt{Target}, \texttt{Content}, and \texttt{Thinking}. The interaction among the modules in the understanding procedure can be expressed as ``I understand the \texttt{Content} of the \texttt{Target} through a way of \texttt{Thinking}". Each module has corresponding story elements as follows.

\begin{itemize}
    \item \texttt{Target}: Character, Object, Place, Conversation, Behavior, Event, Emotion, Commonsense
    \item \texttt{Content}: Identity, Feature, Relationship, Means, Context, Sequence, Causality, Motivation
    \item \texttt{Thinking}: Recall, Recognize, Reasoning
\end{itemize}

The story elements can be assigned on the questions to specify required information, knowledge, or thinking ability by each question. Also, the correct answer rate for each story element is calculated by combining results from the answers. Through this analysis, we can identify the strengths and weaknesses of the video understanding intelligence of each player very specifically.
The set of questions for the VTT can be organized so that CogME attributes appear evenly across the questions. This means that all story elements related to video understanding intelligence can be assessed with a set of questions.

\subsection{The composition of human players}
Human intelligence is not uniform, and there are significant differences in the cognitive aspect by age group according to their developmental stages and individual characteristics. Each person has a different pattern of strengths and weaknesses in their performances and related components of cognition. For example, children are generally good at recall but clumsy in reasoning, while some adults usually focus on the main character in a video clip and ignore surroundings. Therefore, judging the human-likeness of an AI agent can be influenced by the diverse characteristics of the interrogator or the human player. Ultimately, it becomes difficult to generalize the results of the test. To evaluate the human-likeness of an AI agent, VTT conducts tests with a group of human players rather than a single human player. Depending on the characteristics of the human player group, the test can derive more concrete and specified results. For example, if test organizers want to evaluate an AI agent based on age groups, human player groups can be organized based on the developmental stages of Piaget's theory \cite{Piaget1972CogDev}. 
%Participants of a VTT consist of juries and a group of players including an AI agent and human players.

\begin{figure*}[t]
\centering
\includegraphics[width=0.85\textwidth]{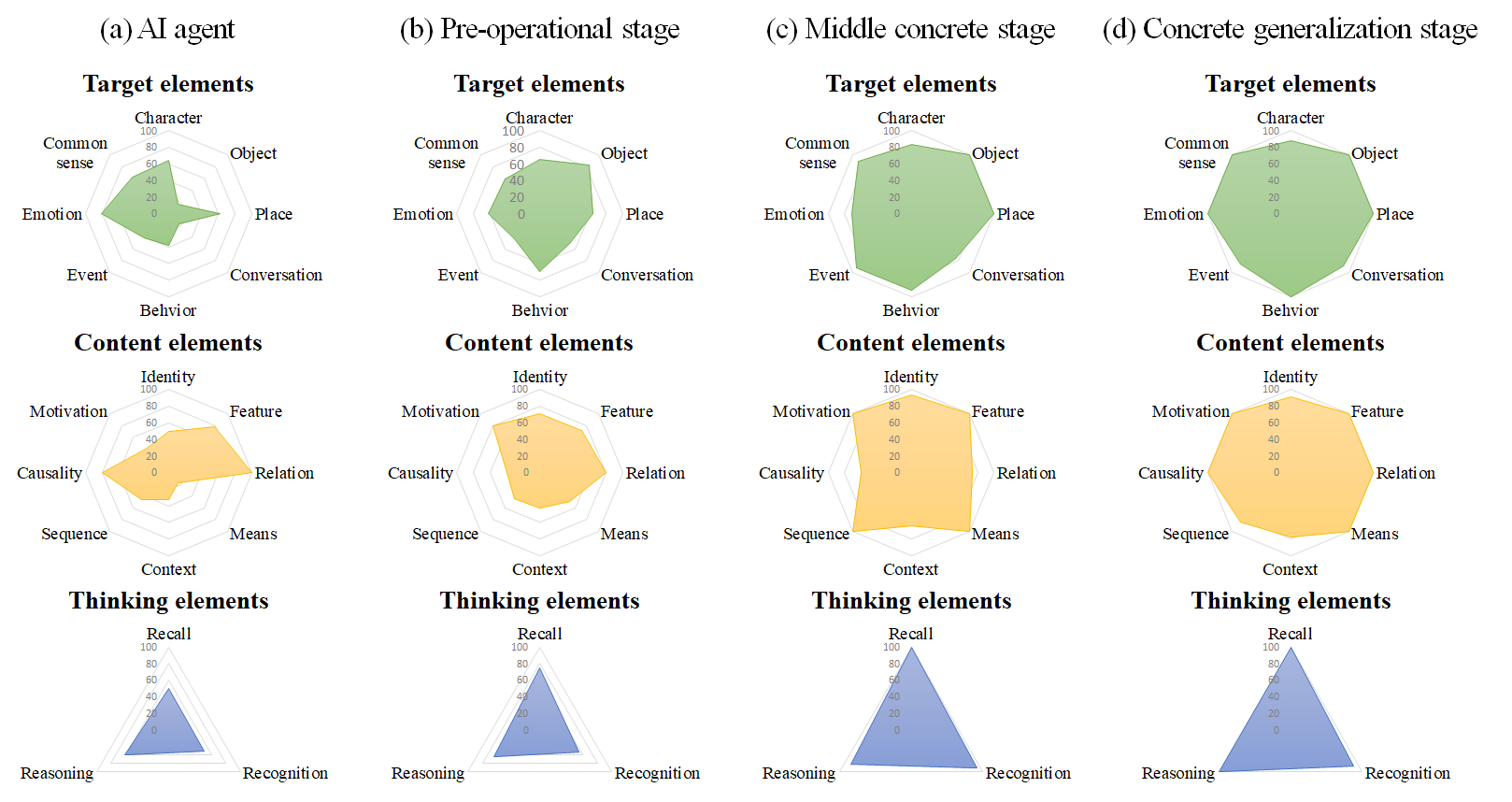}
\caption{The accuracy-based profiles for each player participated in the case study of the Video Turing Test (VTT). To evaluate each player's video story intelligence, we apply CogME based on cognitive process of human.}
\label{fig:analysis}
\end{figure*}

\section{Case Study}
Here, we provide a case study of the VTT to confirm the feasibility and effectiveness of the proposed test. We explain detailed implementations of the case study in this section.

\noindent \textbf{Video selection.  } We utilize shot and scene clips sampled from a Korean popular TV drama ``Another Miss Oh``.

\noindent \textbf{Question selection.  } 
We carefully select 30 questions for the case study considering i) the distribution of story elements, ii) question type (i.e., 15 multiple-choice QA vs. 15 open-ended QA), and iii) associated video length for each question (i.e., 17 QA for shot clip and 13 QA for scene clip).

\noindent \textbf{Human players.  } 
We organize a human player group that consists of four people from the different cognitive developmental stages of humans~\cite{hierarchicalQnA2019heo}. The age of the four players is 4 years (pre-operational stage), 10 years (middle concrete stage), 14 years (concrete generalization stage) and 20+ years (formal stage), respectively.

\noindent \textbf{AI agent.  } 
As an AI agent, we employ two video QA algorithms: \cite{choi2021dramaqa} for multiple-choice QA and \cite{lee2021mounting} for open-ended QA. 

\noindent \textbf{Juries.  } 
As juries for the test, we invite 16 people of ages from the 20s to the 50s and both sexes. Six people of juries are experts for AI and others are non-expert audiences.

\noindent \textbf{Miscellaneous.  } 
We divide 30 questions into 5 rounds of 6 questions. All participants are randomized after each round.
The juries are provided with a judgment sheet to record one's thoughts of each QA. After each round consisting of 6 QAs, the juries have to guess who is the AI, based on the judgment sheet as a cue.
% confidence score에 대한 내용 추가

\section{Results}
%In this section, we present the results of the case study of VTT and analyze the results quantitatively and qualitatively.

\subsection{Question answering and juries' decision} 
First, we present examples of question answering conducted in the case study. In Table \ref{table:qual}, given questions, story elements associated with the questions, and answers generated by players in different developmental stages are described. As shown in the table, our AI agent or few humans fails to answer a given question. At each round after conducting six question and answering pairs, the juries guess who is the AI among the players. As a result, juries perceive the AI around 62.5\% and 75\% at the first and second rounds, respectively. On the contrary, in the third round and fifth round, juries vote one of the human players as an AI around 62.5\% and 87.5\%, respectively. Interestingly, in the fourth round, the juries are quite confused and the voting results were a close match. The juries guessing the AI agent as an AI around 55.6\% (correct guess), and guessing one of the human players as an AI around 45.4\% (incorrect guess). In 4 rounds among a total of 5 rounds, one player receives more than half of the votes with a consistent result\footnote{The juries guess the AI agent as an AI around 48.6\% on average for 5 rounds.}.

\subsection{Analysis on cognitive modules}
We analyze each player's performance based on CogME for specifying their video understanding intelligence. We calculate the correct answer rate (accuracy) for each story element associated with the questions used in the case study. The accuracy-based profiles for each player on story elements in the three cognitive modules (i.e., target, content, thinking) are shown in Figure \ref{fig:analysis}. Here, we highlight two key observations as follows. First, as the developmental stages of human progress, the accuracy of QA across all three modules is gradually improved. This means that some cognitive processes mature in a way that is reflected in the chosen metrics of this test. This result validates the use of these measures to evaluate video understanding intelligence.
%as a person grows, the video understanding intelligence generally tends to mature. 
Second, when comparing the profiles between the AI agent and the human players, we notice that the performance of each component revealed more evenly in human players than the AI agent. Quantitatively, the standard deviation of the accuracy of each story element is 13.78 for a child in the pre-operational stage and 23.13 for the AI agent. In particular, the AI shows deficient performance in Object and Conversation in Target elements, Means and Motivation in Context elements, and Recall in Thinking elements. It is expected for two reasons that the AI agent's profile differs from the previous paper~\cite{shin2021cogme}. This study considers two types of QA, multiple-choice and open-ended, while the previous work use only the former. Furthermore, we employ two distinct video QA algorithms as our AI agents to answer multiple-choice and open-ended QA.

%The first is that this study adopted 50\% full-sentenced open-ended questions, while the previous one used only multiple-choice questions. The second is that two different agents answered in multiple-choice and subjective sentences. 

\section{Discussion and Conclusion}
In this paper, we introduced the Video Turing Test (VTT), a novel measurement to evaluate a human-likeness of video understanding intelligence. We defined a general format and procedure of VTT and conducted a case study to confirm the feasibility and effectiveness of the proposed test. The case study provided new insight into the association of video understanding intelligence between the AI agent and human players from the different developmental stages.

While the case study suggested a new perspective of measurement for video understanding intelligence, we still need to discuss several aspects. First, a 4-years child was included in the test as a human player for comparing the video understanding ability between the AI agent and a human from the pre-operational stage. However, several requirements (e.g., writing as a full sentence, choosing an answer among five answer candidates) were not familiar for the child so that it is hard to interpret the answers as totally reflecting the child's video understanding ability. 
Furthermore, we did not strictly specify the criteria to pass the VTT. The passing criteria can be defined by considering the detailed design of the VTT such as the composition of human players, juries, and the selected question set. As future work, we expect that the additional case study and its interpretation validates the clearness of VTT. For more objective and analytical evaluation for VTT, we plan to conduct various case studies and suggest appropriate guidelines including the composition of participants and the arrangement of question set.

\section{Acknowledgements}
This work was partly supported by the IITP (2015-0-00310-SW.StarLab/20\%, 2017-0-01772-VTT/20\%, 2018-0-00622-RMI/15\%, 2019-0-01371-BabyMind/15\%, 2021-0-02068-AIHub/15\%) grants and the NRF of Korea (2021R1A2C1010970/15\%) grant funded by the Korean government.

%This work was partly supported by the Institute of Information \& Communications Technology Planning \& Evaluation (2015-0-00310-SW.StarLab/20\%, 2017-0-01772-VTT/20\%, 2018-0-00622-RMI/20\%, 2019-0-01371-BabyMind/20\%, 2021-0-02068-AIHub/20\%) grant funded by the Korean government.

\bibliography{aaai21}
\end{document}